# Performance Evaluation of Edge-Directed Interpolation Methods for Images

**Abstract**

Many interpolation methods have been developed for high visual quality, but fail for inability to preserve image structures. Edges carry heavy structural information for detection, determination and classification. Edge-adaptive interpolation approaches become a center of focus. In this paper, performance of four edge-directed interpolation methods comparing with two traditional methods is evaluated on two groups of images. These methods include new edge-directed interpolation (NEDI), edge-guided image interpolation (EGII), iterative curvature-based interpolation (ICBI), directional cubic convolution interpolation (DCCI) and two traditional approaches, bi-linear and bi-cubic. Meanwhile, no parameters are mentioned to measure edge-preserving ability of edge-adaptive interpolation approaches and we proposed two. One evaluates accuracy and the other measures robustness of edge-preservation ability. Performance evaluation is based on six parameters. Objective assessment and visual analysis are illustrated and conclusions are drawn from theoretical backgrounds and practical results.

## 1    Introduction

Images with high resolution and fine details are always admirable and required in many visual tasks. Limited by sensor manufacturing techniques and chip size, post-processing of interpolation techniques are useful and promising [1]. The major advantage of interpolation techniques is that it may cost less and the existing equipments can be utilized. To magnify a region of interest in surveillance, forensic, scientific, medical imaging has been proven useful in many practical cases.

Many interpolation methods for high visual quality have been developed in image processing [1-3], and problems still exist. These problems are highly related to image edges, including the blurring of edges, blocking artifacts in diagonal directions and inability to generate fine details [3]. For the importance of edge-preserving in application fields, edge-adaptive interpolation approaches become a research center of focus [4], and a large number of edge-directed interpolation (EDI) methods have been presented [3-12].

Li [5] proposed a new edge-directed interpolation (NEDI) which takes geometric duality to estimate covariance of targeted high resolution (HR) area from that of local window pixels in low resolution (LR). By the fourth-order linear interpolation, it obtained a HR image of $2^n$ sizewith well-maintained edges.

Its basic assumption is that there are significant correlation between LR image and HR image which is an inadequate approximation and easy to introduce artifacts in high frequency region. And [6] discussed its several problems and analyzed NEDI from window shape, edge pixel handling, error propagation and global brightness invariance. The deep improvement provides better results at the cost of huge computational complexity. After that, [7] proposed a modified version which adopted a modified training window structure to eliminate the predication accumulation problem and extended the covariance matching into multiple directions to suppress the covariance mismatch problem. These modifications can slightly improve image quality with high cost of computation. The understanding of NEDI was reinterpreted as a least-squares estimation method of neighborhood patterns [8], and the paper introduced non-local means as a weighting method to obtain robust improvement used in multi-valued diffusion weighted images. Time cost limits these extended versions of NEDI and their applications in real-time requirement.

Another disadvantage of NEDI is its high computational time cost. Zhang [9] used a directional filtering and data fusion to reduce computational complexity and ringing artifacts (EGII).Chen [10] developed a fast edge-oriented algorithm which partitions digital images into homogeneous areas and edge areas based on preset threshold values. Similarly, [11] detected the presence of edges, and classified the edge-directed interpolation into diagonal process and non-diagonal process. The speeds of [10-11] are competitive to polynomial interpolation, but the area classification is dependent on the preset threshold. Andrea [12] proposed an iterative curvature based interpolation (ICBI) method for real-time application with artifact-free interpolated images. And Zhou [13] introduced edge-adaptive idea into cubic convolution interpolation (DCCI) method and improved image PSNR with reasonable time cost.

The last disadvantage of NEDI may be its $2^n$ integer enlargement factor. In real requirements, interpolation procedure is step by step. From this aspect, many existing EDI approaches are no better than traditional methods.

This paper gives performance evaluation among two traditional interpolation methods, bi-linear and bi-cubic, and four state-of-the-art edge-directed interpolation methods, NEDI [5], EGII [9], ICBI [12] and DCCI [13] based on two groups of standard test images [14]. Objective assessment based on six and two new assessment parameters, and visual analysis is illustrated and conclusions are drawn from theoretical backgrounds and practical results.

The structure of this paper is organized as follows. Section 1 gives an overview based on the analysis of disadvantages of NEDI method. Section 2 defines two parameters from accuracy and robustness to evaluate edge-preserving abilities of interpolation methods. Section 3 describes the state-of-the-art EDI methods. Section 4 introduces experimental materials and evaluation metrics. Section 5 demonstrates performance from quantitative evaluation, and visual quality. And Section 6 discusses applicability of these four EDI methods and future extensions.

## 2   Edge-Preserving Ability

Edges carry heavy structural information for human vision system which leads to detection, classification and determination. Edge detection in general is to preserve the structure properties with reduced amount of data. Canny detector [15] is distinguished from other detectors from three aspects. First, it maximizes the probability to mark real points and reduces the number of non-edge points (good detection). Second, detected edge points are close to the center of edge (good localization), and better than that detected from other edge detectors. Finally, the detected edges are of one pixel width (low spurious response). It has become one of the standard edge detectors for its effectiveness, accuracy and robustness [15-17].

Hereafter, edge is defined restrictedly to edge points on edge map. How to measure the edge-preservation abilities of these EDI methods has not been mentioned. Here, we evaluate it from edge-preserving ratio from rightness and robustness by comparing the interpolated images with corresponding standard images. Canny detector is selected as the tool to extract edge from images, and the threshold of distinguishing edge points is automated settled.

Assuming edge points are extracted by Canny detector, edge points of the standard and interpolated image are $EMs$ and $EMi$. Then we defined two parameters from edge-preserving abilities from accuracy $EPRa$ and robustness $EPRr$ respectively.

$$EPRa = \frac{EMs \cap EMi}{EMs} \qquad (1)$$

$$EPRr = \frac{EMs \cap EMi}{EMs \cup EMi} \qquad (2)$$

## 3. Evaluated Methods

In this section, a brief description of four state-of-the-art EDI methods is depicted. Deep insight into the theories and implementations can be referred to original papers and authors' website [5, 9, 12-13].

The EDI methods share the same procedure, and HR images are filled in three steps. Here, take enlargement factor as $2^1$ for instance, and Figure 1 shows the procedure. First, the HR image is fixed with pixels in LR image from left to right and up to down (dark pixels). Then pixels indexed by two odd values are determined as a weighted average of its four diagonal neighbors (red pixels). Finally, the other pixels (white pixels) are filled with its vertical and horizontal neighbors (red and dark pixels) by the same rule. If the factor is $2^n$, the HR image is obtained with the same procedure by iteration $n$ times.

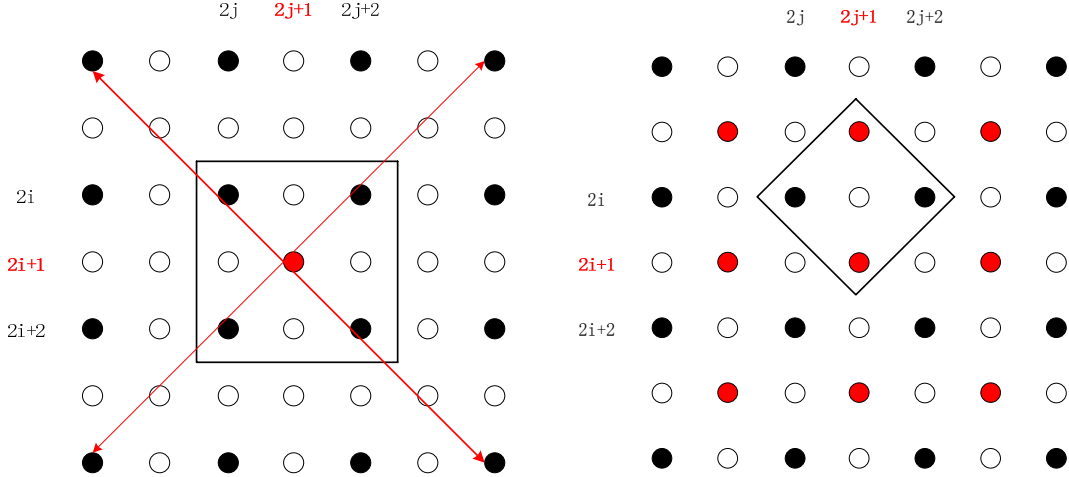

Figure 1 Main step of EDI methods. The emphasis is direction determination of red pixels indexed by two odd values. And categories are 45-degreee diagonal and 135-degree diagonal.

The emphasis is direction determination of pixels between categories of 45-degreee diagonal and 135-degree diagonal, and the main difference for these four methods is the rule of how to weight the four neighbor pixels by determined direction. In NEDI method, no direction determination is considered, and the weights are computed by assuming the local image covariance constant in a large window and at different scales. With this constraint, an over-constrained system of equations can be obtained and solved for the coefficients. This rule applied on homogeneous areas leads to high computational cost [10]. Different from NEDI, the first step of ICBI is to compute local approximations of the second order derivatives along the two diagonal directions using eight pixels (two-arrow red lines), and then assigning the center pixels (red pixels) with the average of the two neighbors in the direction where the derivative is lower. And interpolated values are modified in an iterative procedure trying to minimize an energy function by a greedy strategy. While in the EGII method, the directional estimates, modeled as different noisy measurements of the missing pixels, are fused by the linear minimum mean square-error estimation (LMMSE) technique using the statistics of the two observations sets. And DCCI classifies the direction into much dedicated categories in these main steps. To determine the weights, it takes several natural images to train for acquiring precise weights for final intensity set.

The focus of four methods is the precise estimation of directions. NEDI is from covariance, ICBI is from the second order derivatives, EGII is from statistical analysis of two observation sets, and DCCI is from second order derivatives and parameter are determined by data training. How to utilize the information in LR images and infer the hidden pixel values is still a challenging topic for image interpolation is a severe ill-posed problem.

## 4 Materials and Experiments
### 4.1 Materials

Two groups of standard images are collected. One is 12 digital images from [14], and the other is 12 frames of fetal spine MR images. The size of original digital images are 512×768, and the size of MR images are 360×

320. High resolution is always admirable in many application fields, especially in medical image analysis. Fetal spine MRI is an essential routine for prenatal examination, pregnancy care planning, and postnatal facilitation. It is the first-hand materials to diagnosis and treatment planning for fetus with spine abnormalities. It can provide efficiently complementary assessment of fetal spinal anatomy, especially when suspected malformations occur [18-19].

### 4.2 Metrics

Six metrics are cautiously selected to evaluate the interpolation performance. There are Signal-to-Noise ratio (SNR), Structure Similarity Index (SSIM, [20]), Peak Feature Similarity Index (FSIM, [21]), Mutual Information (MI, [22]), and Time Cost. And then theoretical backgrounds are analyzed from practical interpolated results.

Image quality assessment (IQA) is difficult [23]. SNR and PSNR are from error sensitivity based model and widely used as objective image distortion metrics. And SSIM and FSIM measure structure information maintenance from pixel level and feature level. Mutual information is a basic concept of information theory to measure the statistical dependence and the amount of information. For application requirements, time cost is introduced to measure real-time ability.

### 4.3 Software and Platform

Codes of NEDI, EGII, ICBI and DCCI are from corresponding authors' website [24-27] with no modification. The bi-linear and bi-cubic methods are embedded in MATLAB. These methods are all implemented in MATLAB and running on a PC with Intel (R) Core (TM), i3-2120 CPU @ 3.30 GHZ, 3.29 GHZ, and 1.98 GB DDR RAM on Win7 operation system.

### 4.4 Experiment Design

We evaluate performance from two aspects. The first aspect is from quantitative assessment. For comparison, the standard images are zoomed-in by 50% proportion by simply taking the left-up-corner pixels in each $2\times 2$ pixels (see Eq.3). The second aspect is from visual analysis. We select one image from each standard group with lowest SNR. Then the images are scaled to 50% proportion. The next step is to enlarge them by factor $2^2$. Regions of interest (ROIs) are delineated, and further discussions are based on the analysis of ROI structures and theoretical backgrounds.

Assume an LR image $I_l$ directly down-sampled from an associated HR image $I_h$ through Eq.3.

$$I_l(i,j) = I_h(2i-1, 2j-1), 1 \leq i \leq N, 1 \leq j \leq M \qquad (3)$$

## 5. Results

### Quantitative assessment

Figure2 and Figure3 demonstrate the metrics of two traditional methods and four EDIs from SNR, PSNR, SSIM, FSIM, MI, TC and EPRs for accuracy and robustness. Different methods have different colors. Bi-linear method is in black color of star (*), bi-cubic method is in blue color of left direction (>), NEDI is green color of diameter (◇), EGII is red with dot (.), ICBI is brilliant blue of addition signal (+) and DCII is pink of circle (o).

It is obvious to find that seven IQA parameters of EDI methods are better than traditional methods except time cost. For high complexity of algorithms, most time is consumed on the estimation of pixel categories and weights determinations. Both the fifth image in group one and the second image in group two get worst IQA value in the first five parameters, and they are selected to demonstrate visual quality and detail analysis.

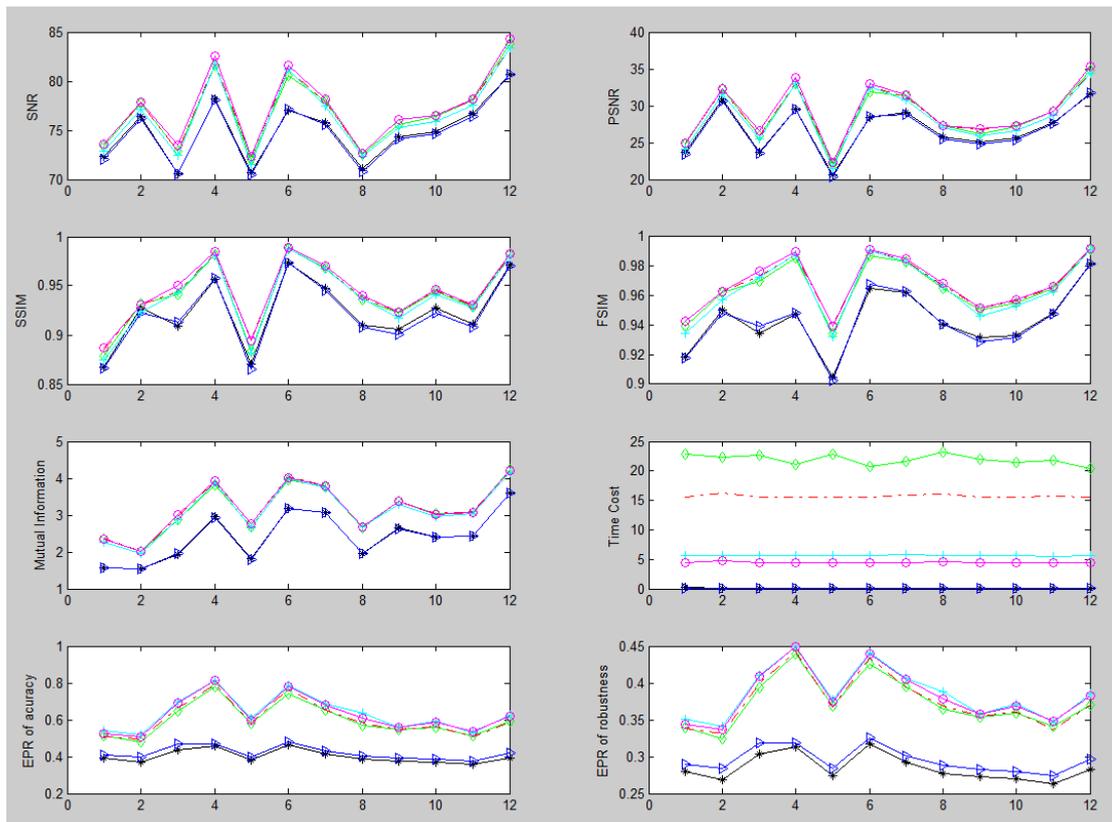

Figure 2 Standard digital images. IQA values of EDI methods are much better than that of traditional methods except high time consumptions, and time consumptions are cautiously concerned in real-time applications.

Unbiased metrics of interpolated digital images are summarized by averaging the scores in Table1 for objective measurement, and the best score is labeled in bold black. From Table 1, it can conclude that: EDI methods are better than traditional methods except time cost. In EDI methods, SNR and PSNR are increased from 1.67dB to 2.60dB, SSIM is from 0.016 to 0.023, FSIM is from 0.022 to 0.026, MI is promoted from 0.60 to 0.76, EPRa is from 0.178 to 0.234 and EPRr is promoted from 0.077 to 0.101, while the time cost increases from 139 times to 932 times. Time consumptions are cautiously concerned in real-time applications. By improved hardware and fast programming, time cost of EDI methods can be dramatically reduced.

Table 1 Unbiased metrics for these six interpolation methods for the first group of digital images

| EDIs / IQA | Bi-linear | Bi-cubic | NEDI | EGII | ICBI | DCCI |
|---|---|---|---|---|---|---|
| SNR | 74.8847 | 74.7169 | 76.8926 | 77.0228 | 76.5585 | **77.2852** |
| PSNR | 26.8021 | 26.6343 | 28.8101 | 28.9402 | 28.4760 | **29.2026** |
| SSIM | 0.9230 | 0.9211 | 0.9408 | 0.9425 | 0.9386 | **0.9441** |
| FSIM | 0.9431 | 0.9430 | 0.9656 | 0.9678 | 0.9649 | **0.9687** |
| MI | 2.4240 | 2.4201 | 3.1508 | 3.1861 | 3.1233 | **3.1871** |
| TC | 0.0320 | **0.0235** | 21.9112 | 15.6736 | 5.6365 | 4.4619 |
| EPRa | 0.3988 | 0.4197 | 0.5984 | 0.6047 | **0.6325** | 0.6251 |
| EPRr | 0.2847 | 0.2952 | 0.3726 | 0.3748 | **0.3855** | 0.3826 |

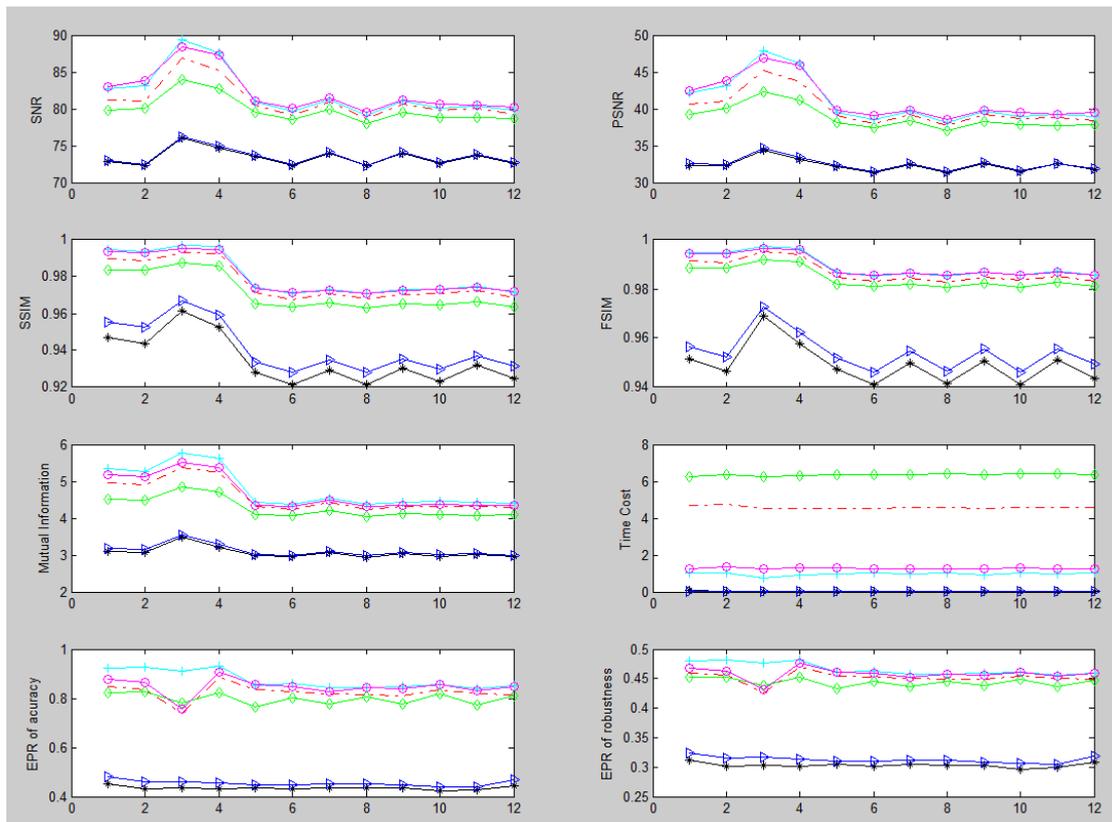

Figure 3 Standard MR Images. IQA values of EDI methods are much better than that of traditional methods except high time consumptions, and time consumptions are cautiously concerned in real-time applications.

Unbiased metrics of interpolated MR images are summarized in Table2, and the best score is labeled in bold black. From Table 2, it can conclude the same conclusion that EDI methods are better than traditional methods except time cost. In these EDI methods, SNR and PSNR are increased from 6.313dB to 8.884dB, SSIM and FSIM are promoted from 0.031 to 0.046, MI is from 1.517 to 1.728, EPRa is increased from 0.343 to 0.439, and ERPr is from 0.131 to 0.163, while the time cost increases from 92 times to 775 times. From edge-generation aspect, traditional interpolation methods generate more details than that of EDI methods for their EPR parameters.

Table 2 Unbiased metrics for these 6 interpolation methods for the second group of MR images

| EDIs \ IQA | Bi-linear | Bi-cubic | NEDI | EGII | ICBI | DCCI |
|---|---|---|---|---|---|---|
| SNR | 73.3713 | 73.5405 | 79.8538 | 81.0679 | 82.0558 | **82.2554** |
| PSNR | 32.3088 | 32.4781 | 38.7914 | 40.0055 | 40.9934 | **41.1930** |
| SSIM | 0.9345 | 0.9410 | 0.9715 | 0.9770 | **0.9803** | 0.9797 |
| FSIM | 0.9492 | 0.9541 | 0.9844 | 0.9869 | **0.9894** | 0.9892 |
| MI | 3.0621 | 3.1145 | 4.2820 | 4.5791 | **4.7897** | 4.6773 |
| TC | 0.0106 | **0.0082** | 6.3543 | 4.5803 | 0.9774 | 1.2787 |
| EPRa | 0.4350 | 0.4553 | 0.7984 | 0.8241 | **0.8738** | 0.8464 |
| EPRr | 0.3031 | 0.3128 | 0.4439 | 0.4516 | **0.4661** | 0.4582 |

From objective analysis, it can conclude that EDI methods are better than the two traditional interpolation methods, bi-linear and bi-cubic except time cost. By proper programming techniques and hardware, these EDIs can dramatically decrease time consumptions, and become feasible to real-time application.

**Visual Quality Analysis**

High visual quality with no error induced is important. We take the 5$^{th}$ image in group one and the 2$^{nd}$ image in group two for visual analysis. The images are zoom-in by 50% proportion, and interpolated with enlargement factor of $2^2$. Meanwhile, two regions of interest (ROI) are delineated to show details in Figure4 and Figure5. The left column is ROI from zoomed-in images, and corresponding interpolated HR images of NEDI, EGII, ICBI and DCCI are shown.

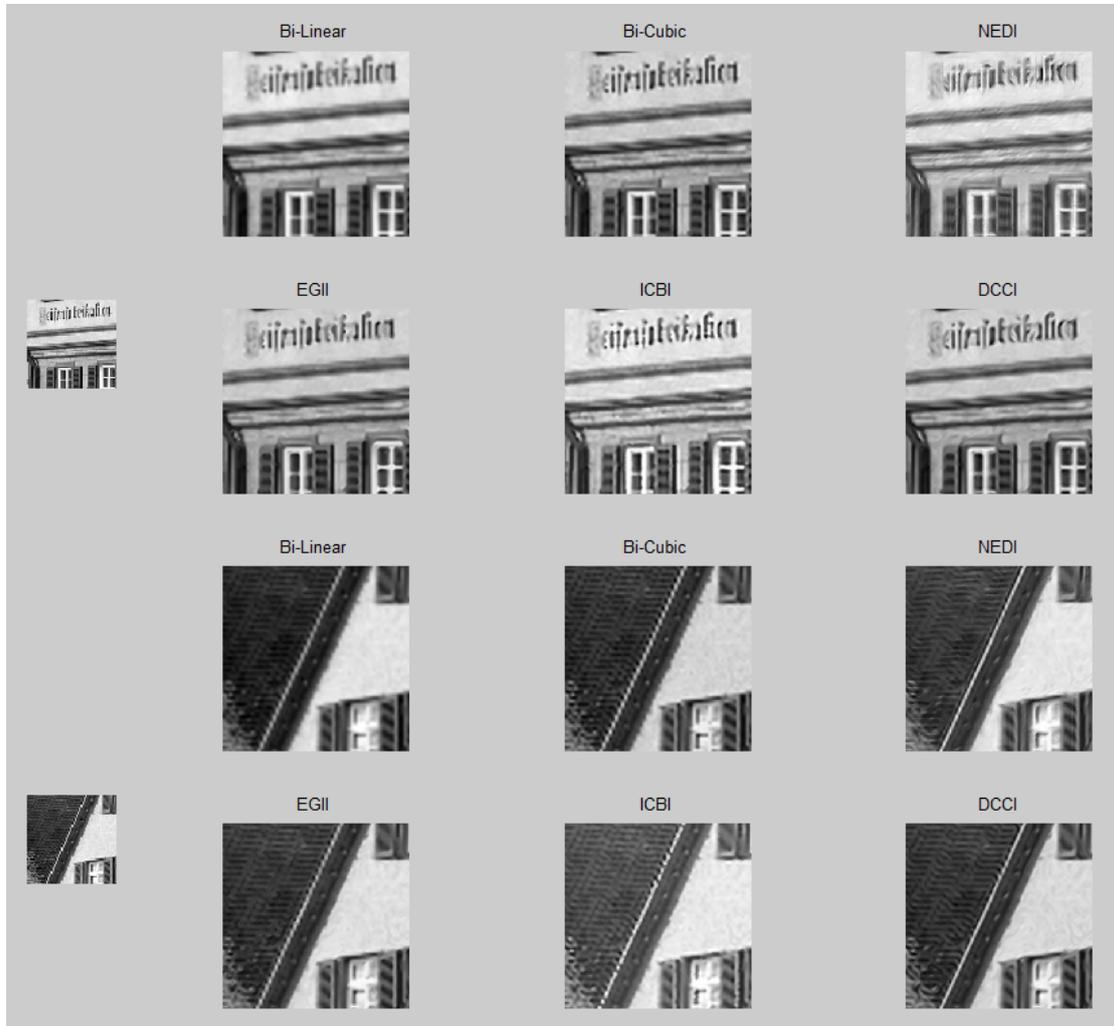

Figure 4 Visual analyses of two ROIs interpolated by different methods. Focuses are characters, windows and tiles.

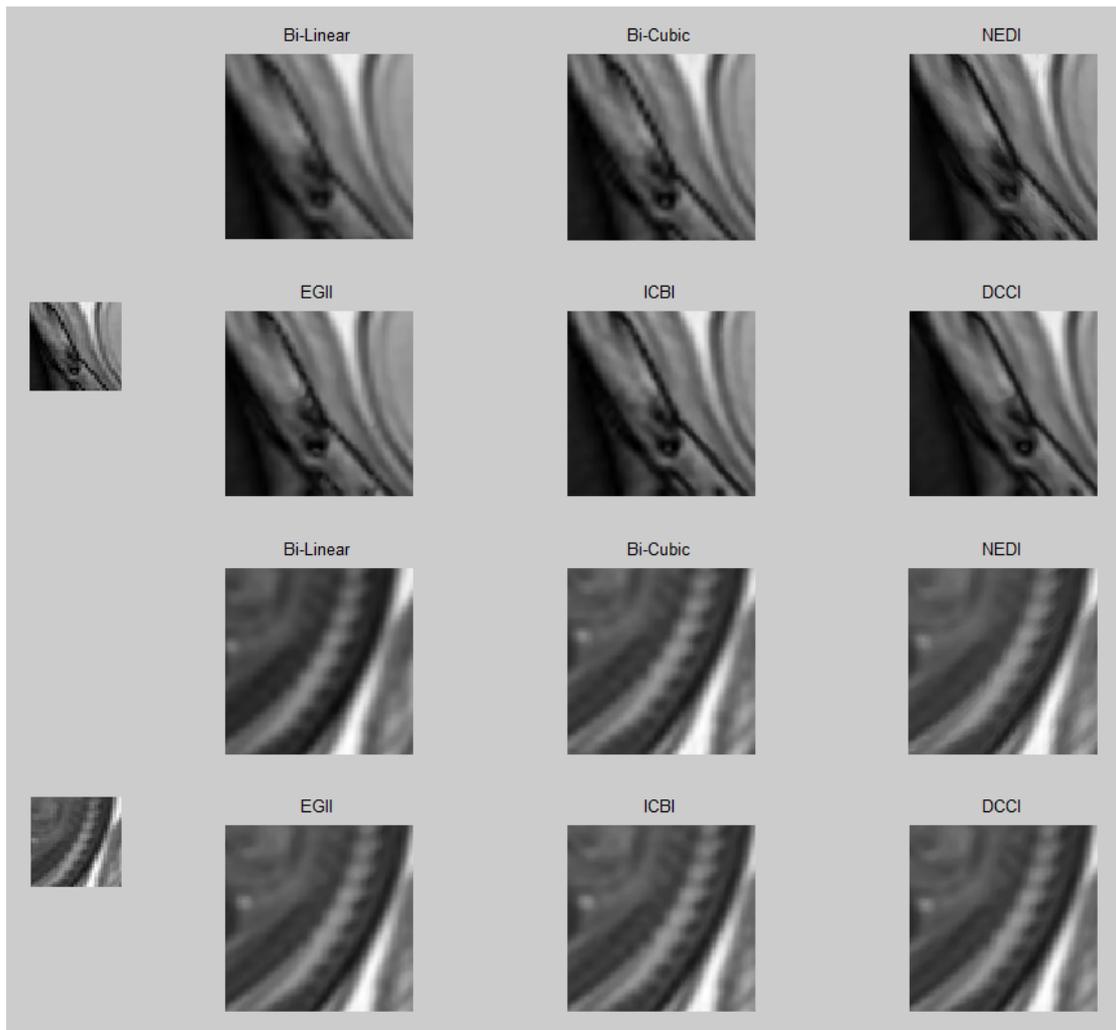

Figure 5 Visual analyses of two ROIs interpolated by different methods. Focuses are line of edges and textures.

Observed from Figure4 and Figure5, we can find that, all interpolation methods obtain HR images with enriched information. Different from bi-linear and bi-cubic method, all EDIs induce unnatural whirlpool-like and blurring textures. It is more obvious in HR images by NEDI.

Different from traditional methods, all EDIs induce unnatural and whirlpool-like textures which are derived from the main steps of estimating orientations of unknown points. This kind of phenomenon is obvious in NEDI for its simple estimations, and restricted by ICBI method. We also conduct performance evaluations from enlargement factor 4. The standard images are down-sampled to 1/16 that is to take one pixel in sixteen to form a new low-resolution image. IQA parameters can reflect no signs of bad visual qualities. If low-resolution images lose its capacity of illustrating the characteristics of scenes or targets, interpolation methods, neither traditional methods nor new EDIs, can't help to recover or generate more meaningful high-resolution images for real applications.

## 5    Conclusion

Image interpolation is intrinsically a severely under-determined inverse problem, and aims to resolve unknown HR image from known LR image, especially the number of unknown HR pixels usually exceeds that of the known LR pixels [19]. Fully interpolation from LR images for high visual quality is still changeling for these existing interpolation methods.

This paper evaluates performance of four EDIs with two traditional methods, bi-linear and bi-cubic, and two parameters to assess edge-preserving ability from accuracy and robustness are proposed. Six interpolation methods are simulated on two groups of images and performance is evaluated from eight parameters and analyzed from visual quality. All EDIs are able to obtain HR images with enriched details, even unnatural textures are induced. Among these four EDIs, ICBI restricts artifacts and recover crisper and de-blurred images. The comparison presented in this paper is not exhaustive and deep investigation can be extent to robustness to noisy images, images of different categories.